\documentclass[10pt,twocolumn]{article}

\usepackage{amsmath}
\usepackage[a4paper,
            top=0.70in,
            bottom=0.75in,
            left=0.70in,
            right=0.70in,
            columnsep=0.25in]{geometry}
\usepackage{times}
\usepackage{microtype}
\usepackage{amsmath,amssymb,amsfonts}
\usepackage{booktabs}
\usepackage{graphicx}
\usepackage{adjustbox}
\usepackage{xcolor}
\usepackage[numbers,sort&compress]{natbib}
\usepackage[colorlinks=true,
            citecolor=blue,
            linkcolor=blue,
            urlcolor=blue]{hyperref}
\usepackage{enumitem}
\setlist{nosep}
\usepackage{graphicx}
\usepackage{siunitx}
\usepackage{xcolor}
\usepackage{caption}
\usepackage{subcaption}
\usepackage{multirow}
\usepackage{array}
\usepackage{float}
\usepackage{url}

\usepackage{hyperref}
\usepackage{cleveref}

\usepackage{hyperref}
\usepackage{cleveref}

\title{From Denoising Dynamics to Endpoint Uncertainty in Diffusion Language Models}

\author{%
  Julio C. Amador D\'iaz L\'opez\thanks{
  SiftyML, LTD\\
  London, United Kingdom \\
  \texttt{julio@siftyml.com}}}

\begin{document}

\maketitle

\begin{abstract}

Predictive uncertainty in stochastic language-model inference can be characterized by repeatedly sampling the same input and measuring the resulting distribution of outputs. However, doing so requires multiple complete generations. Masked diffusion language models expose an additional source of information: the evolution of the output state during denoising. We ask whether convergence observed within diffusion trajectories is systematically associated with uncertainty across their final outputs, which we refer to as endpoints.

We study four masked diffusion language models on subsets of four reasoning benchmarks, collecting 20 stochastic generations per question for a total of 32,000 denoising trajectories. Within each trajectory, we measure within-run concentration, as the area under the curve describing the stable resolution of the eventual output over the denoising trajectory. Across repeated trajectories, we measure the uncertainty of the resulting endpoint distribution.

Stronger within-run concentration is associated with lower endpoint entropy after
accounting for model and dataset ($\beta=-0.222$, $p<10^{-25}$).
This association is robust to restricting the analysis to multi-token outputs
and to reweighting the prevalence of positive-entropy cases. The relationship is strongly asymmetric: high within-run concentration almost exclusively accompanies concentrated endpoint distributions, whereas lower within-run concentration is observed under both concentrated and competing endpoint regimes. This relationship becomes increasingly apparent as denoising proceeds. Most importantly, features from a single denoising run predict whether independent held-out runs will concentrate on a common endpoint or produce competing endpoints.

These results connect dynamics within individual denoising runs to variability across repeated inference. Within-trajectory dynamics contain information about whether independent repetitions of the same inference process will concentrate on a common endpoint or produce competing endpoints. This provides a concrete route towards estimating properties of the endpoint distribution with fewer complete generations.

\end{abstract}

%\newpage

% =======================================================================
\section{Introduction}
\label{sec:introduction}
% =======================================================================

Language-model inference is stochastic. Repeated inference on the same input can produce different outputs, making the distribution of possible outputs or \emph{endpoints} an important source of information about predictive uncertainty. A direct way to characterize this uncertainty is to sample the model repeatedly and measure the resulting endpoint distribution. This approach is conceptually simple but computationally expensive: estimating whether an output is stable, identifying competing endpoints, or measuring their entropy requires multiple complete generations for every input.

Masked diffusion language models expose an additional source of information during inference. Rather than constructing an output through a strictly left-to-right sequence of next-token decisions, they iteratively transform an unresolved output state into a completed sequence. The inference process therefore exposes a temporal trajectory of convergence toward the eventual output. Recent work has begun to characterize these trajectories through token commitment, temporal stability, remasking sensitivity, and other measures of uncertainty during denoising \citep{lu2026temporal,wu2026stabilityweighted,yeom2026answerfirst}. A separate line of work has examined uncertainty in diffusion language models using repeated generations and trajectory-derived signals \citep{vazhentsev2026uncertainty}. Together, these developments raise a more specific cross-scale question: \emph{does the convergence observed within denoising trajectories systematically relate to the distribution of endpoints obtained when the same inference process is repeated?}

This question distinguishes two scales of concentration. \emph{Within} a denoising trajectory, we can measure how much of the eventual output has resolved as inference progresses. \emph{Across} repeated trajectories, we can measure how strongly stochastic inference concentrates on the same final answer. These quantities are different objects: the first describes how the eventual output resolves over the course of a single denoising run, whereas the second describes an empirical distribution over repeated runs. Their relationship is plausible from the structure of diffusion inference, but its empirical strength, temporal evolution, and limits are not determined by that observation alone.

Establishing this relationship matters for uncertainty estimation. Across-run uncertainty is directly observable only after repeating inference. If properties visible during denoising systematically track the endpoint distribution produced across repetitions, then information about that distribution is already present in the trajectory of inference itself. This does not imply that a single trajectory recovers the endpoint distribution. It establishes the prerequisite for asking whether uncertainty that currently requires repeated sampling can eventually be
estimated from one or a small number of denoising trajectories.

We study this relationship in four masked diffusion language models: Dream Base, Dream Instruct, LLaDA-8B Base, and LLaDA-8B Instruct, across subsets of ARC Challenge, MMLU, GSM8K, and BIG-Bench Hard Boolean Expressions. For each
model--dataset--question combination, we collect $K=20$ independent stochastic generations, yielding 32,000 complete denoising trajectories and 1,600 question-level observations.

For each trajectory, we define $S(t)$ as the fraction of the eventual scored output that has stably resolved by normalized denoising time $t$. We summarize the full trajectory by its area under this convergence curve, $c_{\mathrm{within}}$, which measures within-trajectory concentration toward the realized endpoint. Across repeated trajectories, we define $c_{\mathrm{across}}$ as the empirical mass assigned to the modal final answer and use Shannon entropy $H$ to characterize uncertainty in the resulting endpoint distribution. Our primary analysis asks whether variation in $c_{\mathrm{within}}$ is associated with endpoint entropy. Temporal analyses then ask when this relationship becomes visible during denoising.

We obtain four main findings. \emph{First, within-run convergence is systematically associated with across-run endpoint uncertainty. Higher $c_{\mathrm{within}}$ is associated with lower endpoint entropy across multiple robustness specifications}. The association remains statistically significant when analysis is restricted to multi-token outputs, where convergence contains richer trajectory structure than the reveal timing of a single answer token. It also strengthens when positive-entropy observations receive greater weight, indicating that the large mass of unanimous endpoint distributions attenuates rather than generates the pooled association.

\emph{Second, within- and across-run concentration exhibit a strongly asymmetric empirical geometry}. High within-run concentration almost exclusively accompanies high across-run concentration, whereas high across-run concentration frequently occurs despite substantially slower within-run convergence. The two forms of concentration are therefore related but not interchangeable.

\emph{Third, this relationship develops during inference}. The fraction of the eventual output already resolved at intermediate denoising times is associated with endpoint entropy, and the magnitude of this association increases as denoising progresses. Complementary time-to-convergence measures produce the corresponding pattern: later convergence is associated with higher endpoint entropy. The relationship is not uniform across tasks, however; dataset-specific estimates are significant for ARC, BBH, and MMLU but not GSM8K.

\emph{Fourth, denoising dynamics predict endpoint competition in independent held-out runs}. Features from a single trajectory predict whether repeated inference produces competing endpoints substantially better than model and dataset identity alone (AUROC 0.854 versus 0.753). As more trajectories are observed, their final outputs become increasingly informative, but trajectory dynamics provide complementary information: adding temporal features to observed-output features improves AUROC at each evaluated multi-trajectory budget. Thus, information about the endpoint regime is already present within individual denoising runs before that regime is directly observed through repeated inference.

\emph{Finally, convergence does not imply correctness}. Strong within-run convergence and highly concentrated, even unanimous, endpoint distributions can occur around an incorrect answer. The quantities studied here therefore characterize uncertainty in the model's stochastic endpoint distribution rather than correctness of the endpoint.

Taken together, these results connect dynamics within individual denoising runs to the distribution of endpoints produced across repeated inference. Whereas prior work has used denoising trajectories as uncertainty signals or to improve decoding, we show that within-trajectory dynamics predict a distinct object: whether independent repetitions of the same inference process will concentrate on a common endpoint or produce competing endpoints. This links uncertainty visible within a single realization to variability across realizations, and provides a concrete route toward estimating properties of the endpoint distribution with fewer complete generations.

% =======================================================================
\section{Background}
\label{sec:background}
% =======================================================================

\paragraph{Masked Diffusion Language Models.}
\label{sec:background:diffusion}

Masked diffusion language models generate discrete sequences through an iterative denoising process rather than through autoregressive left-to-right factorization. A common formulation defines a forward corruption process that progressively replaces tokens with an absorbing mask state,
\begin{multline}
  q(\mathbf{x}_t \mid \mathbf{x}_0)
  =\\
  \prod_{i=1}^{L}
  \left[
    (1-\alpha_t)\,
    \delta(\mathbf{x}_t^{(i)},[\text{MASK}])
    +
    \alpha_t\,
    \delta(\mathbf{x}_t^{(i)},\mathbf{x}_0^{(i)})
  \right],
\end{multline}
where $\alpha_t$ controls the amount of retained information at diffusion time $t$ and $\mathbf{x}_0$ is the clean sequence
\citep{austin2021d3pm,sahoo2024mdlm,shi2024md4}. The learned reverse process predicts clean token values from partially masked states and progressively reconstructs a complete sequence. Modern masked diffusion language models extend this principle to large-scale generation, including Dream~\citep{ye2025dream} and LLaDA~\citep{nie2025llada}.

At inference time, the model evaluates unresolved positions using the current partially generated sequence as context. A decoding policy then determines which positions are committed or unmasked before the next denoising step. Because masked positions can condition on visible tokens on either side, textual position and resolution order need not coincide. The resulting sequence of intermediate states, $\mathbf{x}_{T} \rightarrow \mathbf{x}_{T-1} \rightarrow \cdots \rightarrow \mathbf{x}_{0}$, therefore exposes an explicit trajectory through which the final output emerges.

\paragraph{Denoising Dynamics and Token Commitment.}
\label{sec:background:trajectory}

The availability of intermediate states has motivated recent work treating denoising time as an object of analysis rather than merely an implementation detail. \citet{lu2026temporal} studies the temporal emergence of information in LLaDA-8B-Base using token commitment, linear probes, confidence and entropy dynamics, and mid-trajectory perturbations. The study finds structured temporal variation in the information available during denoising, including differences in when linguistic properties become recoverable and in uncertainty between tokens that eventually resolve correctly and incorrectly.

Temporal information can also be useful directly for decoding. \citet{wu2026stabilityweighted} argues that token-selection rules based only on confidence at a single denoising step discard information contained in the preceding trajectory. Their Stability-Weighted Decoding method measures temporal instability through changes in predictive distributions across consecutive
steps and uses this history to modify token commitment decisions, improving generation performance on mathematical reasoning and code-generation tasks.

A related question concerns \emph{where} commitment occurs. \citet{yeom2026answerfirst} shows that commitment order in diffusion language models can depart substantially from textual order on reasoning tasks. In unrestricted LLaDA decoding on GSM8K, answer-bearing positions can commit while much of the preceding reasoning text remains unresolved. Restricting long-range
commitment changes both generation behavior and reasoning accuracy. These results show that denoising trajectories contain structure not captured simply by the final output or by the number of completed denoising steps.

Together, this literature establishes that commitment time, commitment order, and temporal stability are informative properties of diffusion inference. The present work builds on this trajectory-level view but studies a different quantity: the cumulative fraction of the eventual output that has stably resolved as inference progresses.

\paragraph{Uncertainty from Diffusion Trajectories.}
\label{sec:background:uq}

Uncertainty in language generation is commonly estimated from either model probabilities or repeated stochastic generations. Sampling-based approaches characterize the distribution over completed outputs directly and underpin methods such as self-consistency~\citep{wang2023selfconsistency,chen2023universal}.
For structured prediction, dispersion across complete sequences provides information that is not available from an isolated token probability \citep{malinin2021uncertainty}. The cost is that estimating such distributions requires multiple complete inference runs.

This trade-off is particularly relevant for diffusion language models, whose iterative trajectories provide additional candidate uncertainty signals. \citet{vazhentsev2026uncertainty} presents a systematic study of uncertainty quantification for large language diffusion models and derives trajectory-based signals from intermediate generations, remasking dynamics, and denoising complexity. Their methods approach stronger sampling-based uncertainty baselines at substantially lower computational cost, demonstrating that the denoising process itself contains information relevant to uncertainty estimation.

The trajectory-based quantities considered in this literature primarily use intermediate dynamics to predict properties such as correctness or hallucination, or to improve the decoding procedure itself. Repeated sampling, by contrast, directly reveals how probability mass is distributed across the endpoints produced by stochastic inference. These constitute different observational scales of the same generative process.

\paragraph{Local and Global Signals in Language-Model Inference.}
\label{sec:background:local-global}

A related line of work considers whether measurements defined at different scales of an inference process provide consistent information about model behavior. In sequential LLM decision systems, local perturbations to input information can propagate into global instability across a sequence of decisions, showing that locally plausible outputs need not compose into stable sequential behavior~\citep{diazlopez2026telemetry}. More directly,
\citet{diazlopez2026confidence} distinguishes local confidence, measured from the probability of a selected answer token, from global confidence, measured from modal-answer frequency under repeated stochastic inference. The two signals are not empirically interchangeable, and their disagreement is associated with sampling instability.

These findings motivate a broader cross-scale question: what can information observed within an individual inference process reveal about behavior that emerges across repeated realizations of that process? Masked diffusion language models provide a particularly rich setting for this question because a single generation exposes an entire denoising trajectory rather than only a token-level confidence signal. This motivates the within--across
distinction considered here.

\paragraph{Within-Trajectory and Across-Trajectory Concentration.}
\label{sec:background:within-across}

The distinction between these two scales motivates the present analysis. Consider one stochastic denoising trajectory. At each normalized inference time $t$, some fraction of the eventual output may already have stably resolved. Tracking this fraction produces a convergence trajectory $S(t)$ describing how the realized endpoint emerges \emph{within} an inference run.

Repeating the complete denoising process produces a second object: an empirical distribution over final endpoints. Its concentration can be summarized by the frequency of the modal answer, while its dispersion can be summarized by Shannon entropy. This is an \emph{across-run} property.

Existing commitment-time analyses ask when particular positions or fractions of an output become stable, while trajectory-based uncertainty methods use intermediate states as diagnostic signals. The question considered here is instead explicitly cross-scale, i.e., whether \text{within-trajectory convergence dynamics} are related to \text{across-trajectory endpoint uncertainty}.

There is no requirement that the two measurements coincide. A set of trajectories may converge relatively slowly within individual runs yet repeatedly terminate at the same endpoint, while stochastic competition among endpoints may be accompanied by different within-run dynamics. The empirical questions are therefore whether these two scales co-vary, when during denoising such a relationship becomes visible, and what information within-run convergence provides about the endpoint distribution observed under repeated inference.

% =======================================================================
\section{Methods}
\label{sec:methods}
% =======================================================================

\paragraph{Models and Benchmarks.}
\label{sec:methods:models}

We evaluate four masked diffusion language models with irreversible token-resolution
dynamics: Dream Base, Dream Instruct, LLaDA-8B Base, and LLaDA-8B Instruct.
%LLaDA2.1-mini is excluded from the primary analysis because its editable block-level generation process permits previously resolved tokens to be revised, making its within-trajectory convergence dynamics non-comparable to the irreversible samplers.
Each model is evaluated on subsets of four benchmarks: ARC-Challenge, MMLU, GSM8K, and the Boolean Expressions task from BIG-Bench Hard (BBH). For each model--dataset combination, we evaluate 100 benchmark items and collect $K=20$
independent stochastic generation trajectories per item. For ARC-Challenge, MMLU, and GSM8K, we randomly shuffle the corresponding test set using seed 42 and select the first 100 items. For BBH, we use the Boolean Expressions task
and select its first 100 items.  The primary analysis therefore contains
$4\ \text{models}
\times 4\ \text{datasets}
\times 100\ \text{questions}=1600$
question-level observations and $1600\times20=32{,}000$ individual denoising trajectories. For each trajectory, the complete denoising history is retained, including the token state at every denoising step and the final generated output.

\paragraph{Within-Trajectory Convergence.}
\label{sec:methods:within}

Let $z_{qr}(t)$ denote the generated sequence state for question $q$, run $r$, at normalized denoising time $t\in[0,1]$, and let $z_{qr}(1)$ denote the final output. For irreversible samplers, a token position is considered \emph{stably resolved} at time $t$ if it has reached its final token value by that point and does not subsequently
change. We define the within-trajectory convergence function
\[
S_{qr}(t)
=
\frac{
\#\{\text{scored positions stably resolved by time }t\}
}{
\#\{\text{scored positions in the final output}\}
}.
\]

Thus, $S_{qr}(t)\in[0,1]$, with larger values indicating that a larger fraction of the eventual output has already been resolved\footnote{This construction is related to the trajectory-based uncertainty signals of \citet{vazhentsev2026uncertainty}, but measures a different quantity. Their metrics use intermediate denoising states to derive uncertainty scores, whereas $S_{qr}(t)$ measures the fraction of the realized final output already
stably resolved at time $t$. We introduce $S_{qr}(t)$ because the present analysis specifically requires a measure of convergence toward the realized output that can be compared with variability across repeated final outputs.}

To summarize the full trajectory with a scalar measure, we compute the area under the
convergence curve:
\[
C^{\mathrm{within}}_{qr}
=
\int_0^1 S_{qr}(t)\,dt.
\]

Question-level within-run concentration is then defined as the mean across the $K=20$
runs:
\[
c_{\mathrm{within},q}
=
\frac{1}{K}
\sum_{r=1}^{K}
C^{\mathrm{within}}_{qr}.
\]

Larger $c_{\mathrm{within}}$ therefore corresponds to earlier and more sustained
resolution of the eventual output over the denoising horizon.

\paragraph{Fixed-Time and Time-to-Convergence Measures.}
\label{sec:methods:temporal}

The AUC measure summarizes the full trajectory but removes explicit temporal
structure. We therefore additionally characterize convergence in two complementary
ways. First, for fixed normalized inference times $u$, we compute $S_q(u)
= \frac{1}{K} \sum_{r=1}^{K}
S_{qr}(u)$, for $u\in\{0.25,0.50,0.75,0.90\}$. %These quantities answer the question: how much of the eventual output has been resolved by a specified point in denoising?

Second, we define time-to-convergence landmarks
$T_{x,qr}
=
\min
\left\{
t:
S_{qr}(t)\ge x
\right\}$, for $x\in\{0.50,0.90,0.95,1.00\}$. Question-level $T_x$ values are averaged across runs. The $S(u)$ and $T_x$ measures therefore provide complementary views of the same trajectory:
\[
S(u)
=
\text{fraction resolved at time }u,
\]
whereas
\[
T_x
=
\text{time required to reach convergence level }x.
\]

\paragraph{Across-Run Endpoint Distribution.}
\label{sec:methods:across}

Repeated stochastic inference induces an empirical distribution over final extracted answers. Throughout, we refer to each final extracted answer as an \emph{endpoint}. For question $q$, let $A_{q1},A_{q2},\ldots,A_{qK}$ denote the final answers obtained from the $K=20$ independent trajectories. The empirical answer distribution is
$\hat P_q(a)
=
\frac{1}{K}
\sum_{r=1}^{K}
\mathbf{1}[A_{qr}=a]$. Across-run concentration is defined as the empirical mass assigned to the modal answer:
$c_{\mathrm{across},q}
=
\max_a \hat P_q(a)$. Across-run uncertainty is measured using Shannon entropy:
$H_q
=
-\sum_a
\hat P_q(a)\log \hat P_q(a)$. We additionally record the number of distinct observed endpoints, whether the endpoint distribution is unanimous, the modal answer, and modal-answer correctness.

\paragraph{Primary Within--Across Association.}
\label{sec:methods:regression}

The primary empirical question is whether variation in within-run convergence is associated with uncertainty across repeated inference. We estimate
\[
H_{qmd}=\alpha_m+\gamma_d+\beta c_{\mathrm{within},qmd}+\epsilon_{qmd},
\]
where $m$ and $d$ index model and dataset. We report specifications with no model or dataset indicators, model indicators only, dataset indicators only, and both model and dataset indicators; the latter is the preferred specification. HC3 heteroskedasticity-robust standard errors are used throughout, with HC1 and standard errors clustered by unique dataset--question pair reported as robustness check.

\paragraph{Temporal Association Analysis.}
\label{sec:methods:temporal-regression}

To determine when within-run convergence becomes informative about across-run uncertainty, we estimate the preferred specification, accounting for model and dataset identity at each normalized inference time,
\[
H_{qmd}=\alpha_m+\gamma_d+\beta(u)S_q(u)+\epsilon_{qmd},
\]
yielding a temporal coefficient profile $\beta(u)$. We separately estimate
\[
H_{qmd}=\alpha_m+\gamma_d+\beta_xT_{x,q}+\epsilon_{qmd},
\]
\noindent where $x\in\{0.50,0.90,0.95,1.00\}$, to relate endpoint entropy to the time required to reach different convergence thresholds.

\paragraph{Joint Within--Across Geometry.}
\label{sec:methods:geometry}

We characterize the joint empirical distribution of within- and across-run concentration by partitioning observations at the descriptive threshold $\tau=0.9$ according to whether $c_{\mathrm{within}}$ and $c_{\mathrm{across}}$ fall above or below $\tau$. Because the marginal distribution of $c_{\mathrm{within}}$ differs substantially across model--dataset cells, we additionally define high within-run concentration relative to the empirical upper quartile of $c_{\mathrm{within}}$.

\paragraph{Competing-Endpoint Subsets.}
\label{sec:methods:competition}

To distinguish the overall association from variation conditional on endpoint competition, we repeat the primary regression for $|\mathcal{A}_q|\ge2$ and $|\mathcal{A}_q|\ge3$, where $\mathcal{A}_q=\{a:\hat P_q(a)>0\}$ is the set of observed endpoints. These specifications ask whether within-run convergence continues to characterize entropy once multiple endpoints have been observed.

\paragraph{Robustness Analyses.}
\label{sec:methods:robustness}

We assess the robustness of the primary $H$--$c_{\mathrm{within}}$ association using HC1 and HC3 standard errors and clustering by benchmark item; restricting the sample to multi-token or single-token outputs, base or instruction-tuned models, Dream or LLaDA-8B models, and individual benchmarks; and excluding model--dataset cells with near-degenerate $c_{\mathrm{within}}$ variance or observations with unanimous endpoint distributions. Because 84\% of observations have zero endpoint entropy, we additionally test whether this mass drives the pooled association using repeated balanced bootstrap samples in which positive-entropy observations constitute 40\%, 50\%, or 60\% of the sample. Balancing is performed within model--dataset cells, and the preferred specification accounting for model and dataset identity is re-estimated over 10,000 bootstrap replicates at each target. We report the bootstrap coefficient
distribution and percentile 95\% interval. All relationships are interpreted as associational.

\paragraph{Predicting Endpoint Competition.}
\label{sec:methods:multiplicity}

We next test whether denoising dynamics from a small number of trajectories
predict endpoint competition in independent held-out runs. For each question
and predictor budget $k\in\{1,2,3,5\}$, we randomly select $k$ of the 20
trajectories for prediction and use the remaining $20-k$ trajectories to
define the held-out target. The primary binary target indicates whether the
held-out trajectories produce more than one distinct endpoint; we additionally
consider a three-class target distinguishing one, two, and at least three
endpoints. Predictor sets comprise model and dataset identity; statistics of
the $k$ observed final outputs; mean $c_{\mathrm{within}}$; temporal features
$S(0.25)$, $S(0.50)$, $S(0.75)$, and $S(0.90)$; and combinations of output
and temporal features. We fit class-balanced logistic classifiers and evaluate
out of sample using five-fold cross-validation grouped by benchmark item, such
that the same item across models cannot appear in both training and test folds.
The trajectory split is repeated 500 times. We report AUROC for binary endpoint
competition and macro-F1 for the three-class task, with 95\% intervals given
by the 2.5th and 97.5th percentiles across trajectory splits.

\paragraph{Correctness Analysis.}
\label{sec:methods:correctness}

We estimate logistic regressions of modal-answer correctness on $c_{\mathrm{within}}$ and $c_{\mathrm{across}}$, separately and jointly, controlling for model and dataset effects, reporting odds ratios for a $0.1$ increase in each measure. We also characterize a ``stable-wrong'' regime as observations with $c_{\mathrm{within}}\ge0.9$ and $c_{\mathrm{across}}\ge0.9$ whose modal answer is incorrect.

% =======================================================================
\section{Results}
\label{sec:results}
% =======================================================================

\paragraph{Within-Run Convergence Is Associated with Endpoint Entropy.}
\label{sec:results:entropy}

Across the full sample, higher $c_{\mathrm{within}}$ is associated with lower endpoint entropy. The unadjusted association is $\hat\beta=-0.212$ (SE $=0.018$). After accounting for model and dataset identity, the estimate remains similar ($\hat\beta=-0.222$, SE $=0.021$, 95\% CI $[-0.263,-0.180]$, $p<10^{-25}$). Accounting for model or dataset identity separately produces similar estimates, and clustering standard errors by benchmark item leaves the result essentially unchanged ($\hat\beta=-0.222$, SE $=0.022$; See \Cref{tab:main} in \Cref{app:additional-results}).

The association remains negative when restricted to multi-token outputs ($\hat\beta=-0.089$, $p=0.036$), and is also present separately for base and instruction-tuned models, Dream and LLaDA-8B, and after excluding near-degenerate AUC cells. Dataset-specific estimates are negative and significant on ARC, BBH, and MMLU, but not on GSM8K ($\hat\beta=-0.026$, $p=0.769$), indicating task heterogeneity.

\paragraph{Within-Run and Across-Run Concentration Occupy Different Regimes.}
\label{sec:results:geometry}

Across the $N=1600$ primary model--dataset--question observations,
$c_{\mathrm{within}}$ is substantially more dispersed than
$c_{\mathrm{across}}$. Mean $c_{\mathrm{within}}$ is 0.612
(median 0.531; IQR 0.497--0.919), whereas mean
$c_{\mathrm{across}}$ is 0.963 and its median is 1.000.
Across-run concentration is strongly ceilinged: 84.0\% of questions
produce the same endpoint in all $K=20$ runs, while 16.0\% ($n=256$)
produce at least two endpoints and 3.5\% ($n=56$) produce at least three.

The joint distribution is strongly asymmetric. At $\tau=0.9$, the
dominant regime is low within-run/high across-run concentration
(1013/1600; 63.3\%), followed by high concentration on both measures
(418; 26.1\%). Low concentration on both measures occurs in 168 cases
(10.5\%), while only one observation (0.1\%) combines high within-run
with low across-run concentration (\Cref{tab:regions}). Using the
empirical top quartile of $c_{\mathrm{within}}$ ($q_{0.75}=0.919$),
all 393 questions have $c_{\mathrm{across}}>0.9$, whereas only 28.0\%
of questions with $c_{\mathrm{across}}>0.9$ belong to the top
within-run quartile. Thus, high within-run concentration almost
exclusively accompanies high across-run concentration, whereas high
across-run concentration frequently occurs despite slower within-run
convergence.

\begin{table}[H]
\centering
\caption{Four-region counts at $\tau=0.9$ ($N=1600$).}
\label{tab:regions}
\begin{tabular}{lrr}
\toprule
Region & $n$ & \% \\
\midrule
within-low / across-high & 1013 & 63.3 \\
within-high / across-high & 418 & 26.1 \\
within-low / across-low & 168 & 10.5 \\
within-high / across-low & 1 & 0.1 \\
\bottomrule
\end{tabular}
\end{table}

\paragraph{The Within--Across Relationship Strengthens Over Denoising Time.}
\label{sec:results:temporal}

The association is detectable at intermediate denoising steps and
strengthens as inference proceeds. After accounting for model and dataset
identity, the coefficient relating $S(u)$ to endpoint entropy increases
in magnitude from $\hat\beta=-0.145$ (SE $=0.016$) at $u=0.25$ to
$-0.183$ (SE $=0.020$) at $u=0.50$, $-0.248$ (SE $=0.022$) at
$u=0.75$, and $-0.275$ (SE $=0.023$) at $u=0.90$. Complementary
time-to-convergence measures show the corresponding pattern: later
convergence is associated with higher endpoint entropy, with
$\hat\beta=0.223$ for $T_{50}$, $0.106$ for $T_{90}$, $0.094$ for
$T_{95}$, and $0.092$ for $T_{100}$ (See \Cref{tab:temporal} in \Cref{app:additional-results}).

\paragraph{Conditional Association Under Endpoint Competition.}
\label{sec:results:competition}

Restricting the analysis to the $n=256$ questions with at least two
observed endpoints changes the composition of the sample substantially.
The uncontrolled association is positive
($\hat\beta=0.186$, $p=0.010$), but is not robust to model and
dataset fixed effects: with both fixed effects, the estimate is negative
but imprecise ($\hat\beta=-0.326$, $p=0.221$). The subset with at
least three endpoints is smaller ($n=56$); for $c_{\mathrm{within}}$,
the model-and-dataset-FE estimate is $\hat\beta=0.293$
($p=0.770$). %Thus, the preferred estimate among observations already known to produce competing endpoints remains unresolved rather than providing a precise conditional relationship.

\paragraph{Robustness to Zero-Entropy Imbalance.}
\label{sec:results:robustness}

The principal negative association is robust to alternative
standard-error estimators, clustering by benchmark item, restrictions
to multi- and single-token outputs, separate base and instruction-tuned
models, separate Dream and LLaDA-8B families, and exclusion of
near-degenerate $c_{\mathrm{within}}$ cells. The multi-token-only
estimate remains statistically significant
($\hat\beta=-0.089$, $p=0.036$), although smaller than the
full-sample estimate; GSM8K is the principal task-level exception.

Because 84.0\% of observations have $H=0$, we additionally test whether
this mass mechanically generates the pooled association using 10,000
balanced bootstrap replicates within model--dataset cells. Under the
50\% positive-entropy target, the achieved share is 43.75\% because
Dream Base/BBH-bool and Dream Instruct/ARC contain no positive-entropy
observations. The mean bootstrap coefficient is $-0.4517$
(median $-0.4517$, SD $=0.0136$), with percentile 95\% bootstrap
interval $[-0.4783,-0.4254]$; all 10,000 coefficient estimates are
negative.

\begin{figure}[t]
    \centering
    \includegraphics[width=0.8\linewidth]{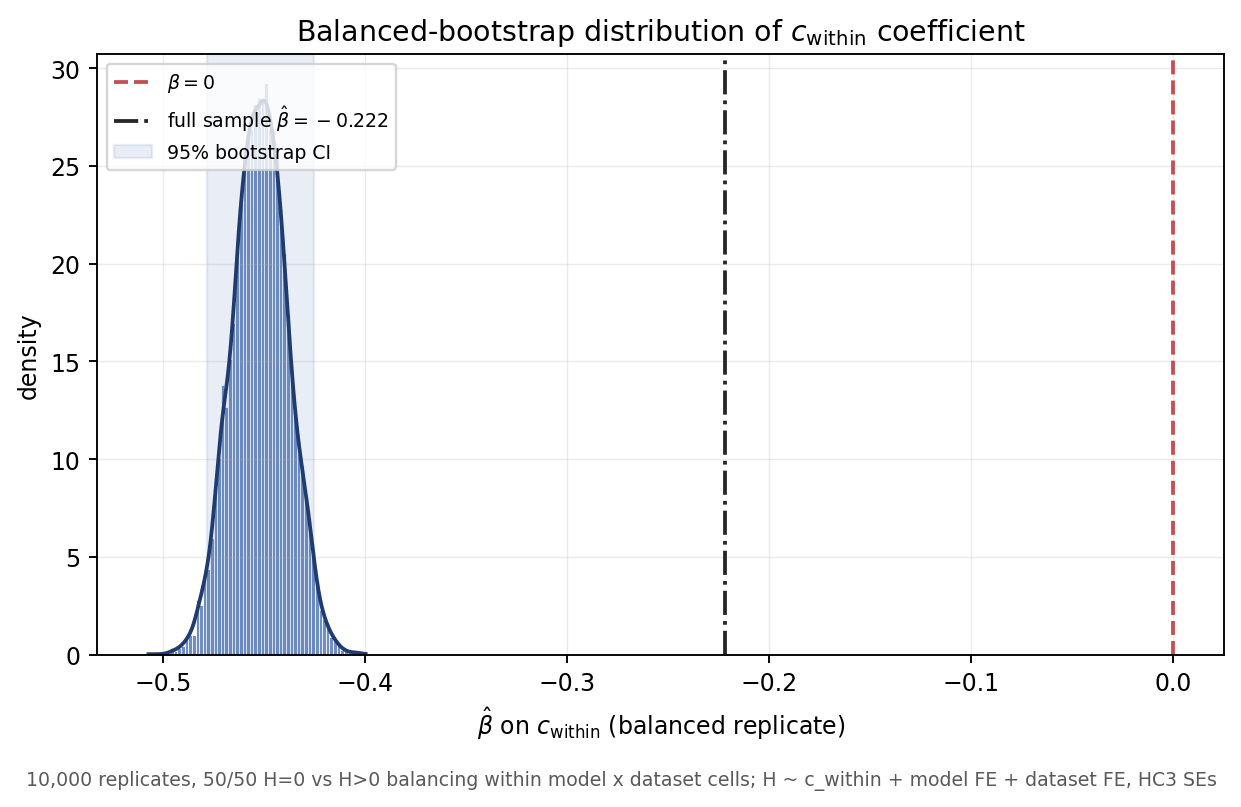}
    \caption{Balanced-bootstrap distribution of the $c_{\mathrm{within}}$
    coefficient under 50/50 balancing of zero- and positive-entropy
    observations within model--dataset cells. The dashed line denotes zero
    and the dash-dotted line the full-sample estimate.}
    \label{fig:balanced-bootstrap}
\end{figure}

The result strengthens as positive-entropy observations receive greater
weight (See \Cref{tab:balanced-bootstrap} and \Cref{fig:balanced-bootstrap} in \Cref{app:additional-results}): mean coefficients are $-0.4094$,
$-0.4517$, and $-0.4835$ under 40\%, 50\%, and 60\% positive-entropy
targets, respectively. Thus, the negative pooled association is not an
artifact of the 84\% zero-entropy mass; under these deliberately
reweighted designs, greater weight on positive-entropy observations
moves the coefficient further from zero. These bootstrap intervals
describe resampling variability under fixed balancing and are not
population confidence intervals.

\paragraph{A Single Denoising Trajectory Predicts Endpoint Competition.}
\label{sec:results:multiplicity}

We next test whether denoising dynamics from a small number of trajectories
predict whether independent held-out runs produce competing endpoints.
Using a single trajectory, temporal features predict endpoint competition
with an AUROC of 0.854 (95\% interval [0.850, 0.857]), compared with
0.753 [0.750, 0.758] using model and dataset identity alone. Macro-F1
similarly increases from 0.407 [0.402, 0.410] to
0.513 [0.499, 0.525]. Because a single observed final output contains
no information about across-run multiplicity by construction, this result
shows that the denoising trajectory contains predictive information about
the outcomes of independent repeated inference.

\begin{figure}[t]
    \centering
    \includegraphics[width=\textwidth]{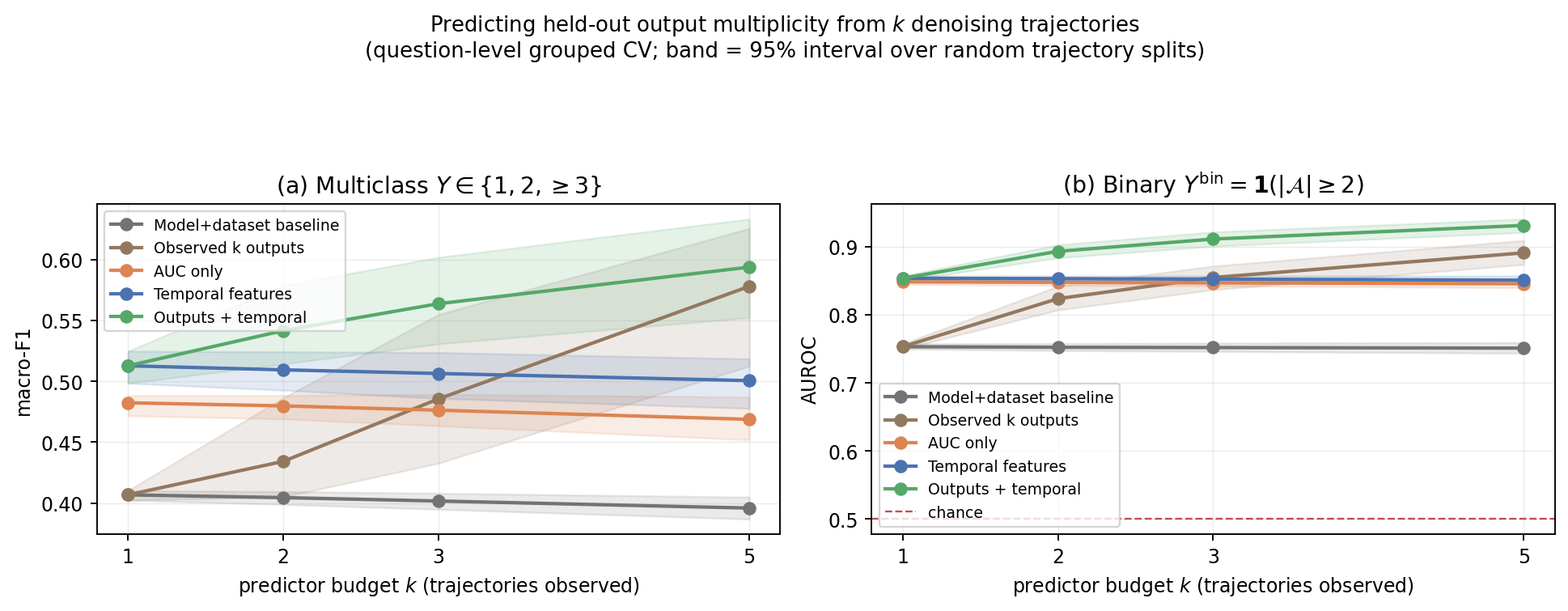}
    \caption{Held-out endpoint-multiplicity prediction as a function of the
    number $k$ of observed denoising trajectories. Bands show 95\% intervals
    across random trajectory splits.}
    \label{fig:endpoint-prediction}
\end{figure}

The predictive value of trajectory dynamics remains approximately stable
as the number of observed trajectories increases, with temporal-feature
AUROC between 0.851 and 0.854 for $k\in\{1,2,3,5\}$. In contrast,
prediction from the observed final outputs improves with repeated sampling,
reaching AUROC 0.891 at $k=5$. The two sources of information are
complementary: adding temporal features to observed-output features improves
AUROC by 0.070, 0.057, and 0.040 at $k=2$, $3$, and $5$, respectively,
with all 95\% intervals excluding zero. Thus, within-trajectory dynamics
provide information about subsequent endpoint competition beyond that
available from a small sample of final outputs (See \Cref{fig:endpoint-prediction} in \Cref{app:additional-results}).

\paragraph{Endpoint Concentration Does Not Imply Correctness.}
\label{sec:results:correctness}

Both concentration measures are positively associated with correctness.
In separate model- and dataset-controlled logistic regressions, a 0.1 increase
in $c_{\mathrm{within}}$ corresponds to an odds ratio of 1.187
(95\% CI [1.134, 1.243], $p=1.6\times10^{-13}$), while the
corresponding odds ratio for $c_{\mathrm{across}}$ is 1.389
(95\% CI [1.235, 1.562], $p=3.9\times10^{-8}$). When both are
included jointly, each remains statistically significant, with odds
ratios of 1.157 for $c_{\mathrm{within}}$ and 1.272 for
$c_{\mathrm{across}}$.

Strong concentration nevertheless does not imply correctness. At
$\tau=0.9$, 78 questions exhibit both high within-run and high
across-run concentration while having an incorrect modal answer,
comprising 4.9\% of the full sample and 13.4\% of all incorrect
observations. Among these stable-wrong cases, 98.7\% are unanimous
across all 20 runs, and 12 benchmark items exhibit the same failure
in at least two primary models.

% =======================================================================
\section{Discussion}
\label{sec:discussion}
% =======================================================================

The results connect two observational scales of diffusion inference. Within a single denoising run, the eventual output progressively resolves from an initially uncertain state. Across repeated runs, stochastic inference produces a distribution over final endpoints. These are distinct objects, but the experiments show that they are systematically related: stronger convergence within individual runs is associated with greater concentration across runs, and information about this across-run regime becomes increasingly visible as denoising proceeds.

\paragraph{Within-run convergence characterizes the endpoint regime.}
\label{sec:discussion:within-across}

The joint distribution reveals a strongly asymmetric relationship. High $c_{\mathrm{within}}$ almost exclusively accompanies high $c_{\mathrm{across}}$, whereas high $c_{\mathrm{across}}$ frequently occurs alongside lower within-run concentration. Repeated trajectories can therefore concentrate on the same endpoint even when within-run concentration is relatively low, but high within-run concentration is rarely observed when repeated inference
produces a dispersed endpoint distribution.

The entropy analysis provides a continuous counterpart to this geometry. Higher $c_{\mathrm{within}}$ is associated with lower endpoint entropy, and the association persists across the principal robustness analyses. Importantly,
reweighting the sample toward positive-entropy observations strengthens rather than weakens the relationship. The large mass of unanimous endpoint distributions therefore attenuates, rather than generates, the pooled association.

\paragraph{The relationship develops during inference.}
\label{sec:discussion:temporal}

The connection between these two scales is already visible before inference terminates. Greater resolved mass at intermediate denoising times is associated with lower endpoint entropy, and the magnitude of this association increases as denoising progresses. Complementary time-to-convergence measures show the corresponding pattern: later convergence accompanies greater endpoint uncertainty.

This temporal structure matters because the final endpoint alone does not contain the path by which it was reached. Two runs may terminate at the same answer while differing substantially in when and how strongly that answer resolved during denoising. The results show that this intermediate trajectory contains information about a property that is defined only across repeated runs: the concentration of the resulting endpoint distribution.

\paragraph{Within-run dynamics predict across-run competition.}
\label{sec:discussion:prediction}

The held-out prediction experiment makes this connection explicit. Features from a single denoising trajectory predict whether independent repetitions of the same inference process will produce a common endpoint or competing endpoints, substantially improving over model and dataset identity alone. As additional runs become available, their observed final outputs become increasingly informative, but trajectory dynamics remain complementary: combining temporal and observed-output features improves prediction beyond the observed outputs alone.

This result places the present findings in a broader trajectory-based uncertainty literature. Prior work has shown that intermediate denoising states can provide useful uncertainty signals or improve decoding. The quantity predicted here is different: it is a property of the distribution generated across independent realizations of inference. The denoising path within one realization therefore contains information about variability across realizations.

This distinction also gives the result a direct practical interpretation. Repeated sampling remains necessary to observe the empirical endpoint distribution itself, but some information obtained from those repetitions is already available within individual denoising runs. Trajectory dynamics may therefore complement repeated sampling when estimating whether inference is concentrated or competitive, potentially reducing the number of complete generations required to characterize stochastic model behavior.

Concentration should nevertheless not be conflated with correctness. Highly concentrated and even unanimous endpoint distributions can converge on an incorrect answer. The quantities studied here characterize uncertainty in the model's stochastic endpoint distribution rather than calibration to ground truth.

\paragraph{Limitations.}
\label{sec:discussion:limitations}

The primary analysis covers four irreversible masked-diffusion models from two model families. LLaDA2.1-mini, for example, is excluded because resolved positions can subsequently be edited, changing the interpretation of stable resolution; whether the same within--across relationship extends to editable or continuous diffusion models remains open. The relationship is also task-dependent, with GSM8K providing the principal exception to the cross-task pattern.

Output geometry affects $c_{\mathrm{within}}$. For single-token irreversible outputs, it is algebraically determined by answer-token reveal time. Although the association remains significant for multi-token outputs, its magnitude is substantially smaller. The strength of the observed relationship should therefore not be assumed invariant to output structure or denoising schedule.

Across-run quantities are estimated from $K=20$ trajectories and remain finite-sample approximations to the underlying endpoint distribution. The competing-endpoint subset is also substantially smaller than the full sample, limiting precision when characterizing dispersion conditional on competition. Finally, the association analyses are observational: they establish systematic covariation between within-run convergence and across-run endpoint uncertainty, not that faster convergence causes lower endpoint uncertainty.

Taken together, the results show that the trajectory of a single denoising run and the distribution produced across repeated runs are linked at two different scales of the same stochastic inference process. Within-run dynamics do more than correlate with endpoint uncertainty: they predict whether independent repetitions will concentrate on a common endpoint or produce competing endpoints. This cross-scale relationship provides a basis for uncertainty methods that use information exposed during denoising to characterize across-run variability with fewer complete generations.

\section{AI Disclosure}
We used generative AI tools to assist with code development and analysis, literature-search support, and manuscript editing. AI-assisted code and analyses were reviewed and validated against the underlying data and experimental outputs, and literature references were manually verified against their original sources. Generative AI was not used to generate experimental data. We take responsibility for the final content of this work, including text, claims, code, analyses, and other artifacts produced with the aid of generative AI.

% -----------------------------------------------------------------------
%  Acknowledgements
% -----------------------------------------------------------------------
% (Omitted for anonymous submission.)

% -----------------------------------------------------------------------
%  References
% -----------------------------------------------------------------------
\bibliographystyle{plainnat}
\bibliography{references}

\newpage
\appendix

\section{Supplementary Material}
\label{app:additional-results}

This supplement reports a descriptive analysis of token
resolution order, followed by additional estimates supporting the main
within--across analysis.

\subsection{Token Resolution Order}
\label{sec:methods:resolution-order}

We additionally examine whether strong within-run convergence exhibits a
systematic positional structure within the generated output.

\paragraph{Analysis.}
For each scored token position $j$ in trajectory $(q,r)$, we define its
stable resolution time as
\[
\tau_{qrj}=\min\{t:z_{qrj}(t)=z_{qrj}(1)\ \text{and unchanged thereafter}\}.
\]
Resolution-order analyses are restricted to multi-token outputs. To avoid
comparing absolute $c_{\mathrm{within}}$ values across cells with different
trajectory mechanics, high-convergence questions are defined as the upper
quartile within each model--dataset cell. Within each cell, we report the
Spearman correlation between token position and resolution time, the
probability that the final token resolves first, and the fraction of
earliest-resolution mass in the first and final positional quartiles.

\paragraph{Resolution Order Is Predominantly Left-to-Right.}
\label{sec:results:resolve-order}

Among high-$c_{\mathrm{within}}$ multi-token trajectories, most
model--dataset cells exhibit a strong positive association between token
position and resolution time. Spearman correlations exceed 0.9 in several
cells, including Dream Base on BBH ($\rho=0.980$) and GSM8K
($\rho=0.999$), Dream Instruct on GSM8K ($\rho=0.999$), and LLaDA-8B
Base on ARC ($\rho=0.935$), GSM8K ($\rho=0.988$), and MMLU
($\rho=0.951$). The pattern is not universal: Dream Base on MMLU shows
weaker ordering ($\rho=0.325$), while LLaDA-8B Instruct on GSM8K shows
the clearest departure ($\rho=0.407$), with 33.3\% of earliest-resolution
mass in the final positional quartile. Resolution order is therefore
predominantly left-to-right among high-convergence trajectories, but its
strength is strongly model- and task-dependent.

\subsection{Additional Within--Across Results}
\label{app:within-across}

Table~\ref{tab:main} reports the primary association under alternative
specifications accounting for model and dataset identity.

\begin{table}[H]
\centering
\caption{Association between $c_{\mathrm{within}}$ and endpoint entropy
when accounting for model and dataset identity.}
\label{tab:main}
\small
\begin{tabular}{lrrrr}
\toprule
Controls & $\hat\beta$ & SE & 95\% CI & $p$ \\
\midrule
None & -0.212 & 0.018 & [-0.246,-0.178] & $<10^{-25}$ \\
Model                     & -0.197 & 0.020 & [-0.237,-0.157] & $5\times10^{-22}$ \\
Dataset                   & -0.229 & 0.018 & [-0.265,-0.193] & $<10^{-25}$ \\
Both         & -0.222 & 0.021 & [-0.263,-0.180] & $<10^{-25}$ \\
\bottomrule
\end{tabular}
\end{table}

Table~\ref{tab:temporal} reports the corresponding temporal estimates
using intermediate resolved mass and time-to-convergence measures.

\begin{table}[H]
\centering
\caption{Temporal association with endpoint entropy after accounting for
model and dataset identity.}
\label{tab:temporal}
\small
\begin{tabular}{lrrr}
\toprule
Predictor & $\hat\beta$ & SE & $p$ \\
\midrule
$S(0.25)$ & -0.145 & 0.016 & $<10^{-18}$ \\
$S(0.50)$ & -0.183 & 0.020 & $1.5\times10^{-19}$ \\
$S(0.75)$ & -0.248 & 0.022 & $<10^{-25}$ \\
$S(0.90)$ & -0.275 & 0.023 & $<10^{-25}$ \\
\midrule
$T_{50}$  &  0.223 & 0.021 & $<10^{-25}$ \\
$T_{90}$  &  0.106 & 0.018 & $3.0\times10^{-9}$ \\
$T_{95}$  &  0.094 & 0.018 & $6.9\times10^{-8}$ \\
$T_{100}$ &  0.092 & 0.018 & $1.8\times10^{-7}$ \\
\bottomrule
\end{tabular}
\end{table}

Finally, Table~\ref{tab:balanced-bootstrap} reports the balanced-bootstrap
analysis used to assess sensitivity to the large zero-entropy mass.

\begin{table}[H]
\centering
\caption{Balanced-bootstrap robustness to zero-entropy imbalance.
Each target uses 10,000 replicates while accounting for model and
dataset identity. First column $H>0$ refers to the target for $H>0$ share, and $\beta<0$ refers to the $P(\beta)<0$.}
\label{tab:balanced-bootstrap}
\small
\begin{tabular}{lrrrr}
\toprule
$H>0$ & Mean $\beta$ & Median $\beta$ &
95\% bootstrap int & $\beta < 0$ \\
\midrule
40\% & -0.4094 & -0.4092 & [-0.4350,-0.3845] & 1.000 \\
50\% & -0.4517 & -0.4517 & [-0.4783,-0.4254] & 1.000 \\
60\% & -0.4835 & -0.4836 & [-0.5115,-0.4545] & 1.000 \\
\bottomrule
\end{tabular}
\end{table}

%\newpage
%\input{checklist.tex}

\end{document}